%% file: main.tex
\pdfoutput=1
\documentclass{article}
\PassOptionsToPackage{numbers}{natbib}

\usepackage[preprint]{neurips_data_2024}

\usepackage[utf8]{inputenc} %
\usepackage[T1]{fontenc}    %
\usepackage{hyperref}       %
\usepackage{url}            %
\usepackage{booktabs}       %
\usepackage{amsfonts}       %
\usepackage{nicefrac}       %
\usepackage{microtype}      %
\usepackage{xcolor}         %
\usepackage{amsmath} 

\usepackage{wrapfig}
\usepackage{tikz}
\usepackage{enumitem}
\usepackage{array}
\usepackage{multirow}
\usepackage{colortbl}
\usepackage{booktabs}
\usepackage{pythonhighlight}
\usepackage{listings}
\usepackage{tcolorbox}
\usepackage{graphicx}

\definecolor{codegreen}{rgb}{0,0.6,0}
\definecolor{codegray}{rgb}{0.5,0.5,0.5}
\definecolor{codepurple}{rgb}{0.58,0,0.82}
\definecolor{backcolour}{rgb}{0.95,0.95,0.92}

\lstdefinestyle{mystyle}{
    backgroundcolor=\color{backcolour},  
    commentstyle=\color{codegreen},
    keywordstyle=\color{magenta},
    stringstyle=\color{codepurple},
    basicstyle=\ttfamily\footnotesize,
    breakatwhitespace=false,         
    breaklines=true,                 
    captionpos=b,                    
    keepspaces=true,                 
    showspaces=false,                
    showstringspaces=false,
    showtabs=false,                  
    tabsize=2
}

\definecolor{keycolor}{rgb}{0,0,0.8}    %
\definecolor{stringcolor}{rgb}{0.5,0,0} %
\definecolor{numbercolor}{rgb}{0.5,0,0.5} %

\definecolor{verylightgray}{rgb}{0.9,0.9,0.9}

\lstdefinelanguage{json}{
    basicstyle=\normalfont\ttfamily,
    showstringspaces=false,
    breaklines=true,
    frame=lines,
    backgroundcolor=\color{verylightgray},
    literate=
     *{0}{{{\color{numbercolor}0}}}{1}
      {1}{{{\color{numbercolor}1}}}{1}
      {2}{{{\color{numbercolor}2}}}{1}
      {3}{{{\color{numbercolor}3}}}{1}
      {4}{{{\color{numbercolor}4}}}{1}
      {5}{{{\color{numbercolor}5}}}{1}
      {6}{{{\color{numbercolor}6}}}{1}
      {7}{{{\color{numbercolor}7}}}{1}
      {8}{{{\color{numbercolor}8}}}{1}
      {9}{{{\color{numbercolor}9}}}{1}
      {:}{{{\color{keycolor}{:}}}}{1}
      {,}{{{\color{keycolor}{,}}}}{1}
      {\{}{{{\color{keycolor}{\{}}}}{1}
      {\}}{{{\color{keycolor}{\}}}}}{1}
      {[}{{{\color{keycolor}{[}}}}{1}
      {]}{{{\color{keycolor}{]}}}}{1}
      {"}{{{\color{stringcolor}{"}}}}{1},
}

\definecolor{lightgray}{rgb}{0.94,0.95,0.95}
\lstdefinelanguage{json2}{
    basicstyle=\normalfont\fontfamily{pcr}\selectfont,
    showstringspaces=false,
    breaklines=true,
    frame=lines,
    backgroundcolor=\color{lightgray},
    literate=
     *{0}{{{\color{numbercolor}0}}}{1}
      {1}{{{\color{numbercolor}1}}}{1}
      {2}{{{\color{numbercolor}2}}}{1}
      {3}{{{\color{numbercolor}3}}}{1}
      {4}{{{\color{numbercolor}4}}}{1}
      {5}{{{\color{numbercolor}5}}}{1}
      {6}{{{\color{numbercolor}6}}}{1}
      {7}{{{\color{numbercolor}7}}}{1}
      {8}{{{\color{numbercolor}8}}}{1}
      {9}{{{\color{numbercolor}9}}}{1}
      {:}{{{\color{keycolor}{:}}}}{1}
      {,}{{{\color{keycolor}{,}}}}{1}
      {\{}{{{\color{keycolor}{\{}}}}{1}
      {\}}{{{\color{keycolor}{\}}}}}{1}
      {[}{{{\color{keycolor}{[}}}}{1}
      {]}{{{\color{keycolor}{]}}}}{1}
      {"}{{{\color{stringcolor}{"}}}}{1},
}

\definecolor{mygray}{rgb}{0.95, 0.95, 0.95}
\definecolor{myblue}{rgb}{0.41, 0.50, 0.57}

\tcbuselibrary{listings,skins,breakable}

\tcbset{
    enhanced,
    breakable,
    colback=mygray, %
    colframe=myblue, %
    boxrule=0.2mm, %
    arc=1mm, %
    top=10pt,
    bottom=10pt,
    left=10pt,
    right=10pt,
    listing only,
    listing options={
        basicstyle=\ttfamily\small, %
        language=Java, %
    },
}

\newcommand*{\affaddr}[1]{#1} %

\title{APIGen: \underline{A}utomated \underline{PI}peline for \underline{Gen}erating Verifiable and Diverse Function-Calling Datasets}

\author{
\text{Zuxin~Liu},
\text{Thai~Hoang},
\text{Jianguo~Zhang},
\text{Ming~Zhu},
\text{Tian~Lan}, 
\textbf{Shirley~Kokane},
\textbf{Juntao~Tan}, 
\\ 
\textbf{Weiran~Yao},
\textbf{Zhiwei~Liu}, 
\textbf{Yihao~Feng}, 
\textbf{Rithesh Murthy}, 
\textbf{Liangwei~Yang}, \\
\textbf{Silvio Savarese},
\textbf{Juan Carlos Niebles},
\textbf{Huan Wang},
\textbf{Shelby Heinecke},
\textbf{Caiming Xiong} \\
\affaddr{Salesforce AI Research, USA} \\
\affaddr{\{zuxin.liu, thai.hoang, jianguozhang\}@salesforce.com}
}

\begin{document}

\maketitle

\begin{abstract}
The advancement of function-calling agent models requires diverse, reliable, and high-quality datasets. 
This paper presents APIGen, an automated data generation pipeline designed to synthesize verifiable high-quality datasets for function-calling applications. We leverage APIGen and collect 3,673 executable APIs across 21 different categories to generate diverse function-calling datasets in a scalable and structured manner. Each data in our dataset is verified through three hierarchical stages: format checking, actual function executions, and semantic verification, ensuring its reliability and correctness. 
We demonstrate that models trained with our curated datasets, even with only 7B parameters, can achieve state-of-the-art performance on the Berkeley Function-Calling Benchmark, outperforming multiple GPT-4 models. Moreover, our 1B model achieves exceptional performance, surpassing GPT-3.5-Turbo and Claude-3 Haiku. We release a dataset containing 60,000 high-quality entries, aiming to advance the field of function-calling agent domains. The dataset is available on Huggingface~\footnote{\url{https://huggingface.co/datasets/Salesforce/xlam-function-calling-60k}} and the project homepage~\footnote{\url{https://apigen-pipeline.github.io/}}.

\end{abstract}

\section{Introduction}
\input{sec1_intro_new}

\section{Related Work}
\input{sec2_related_new}

\section{APIGen Framework}
\input{sec3_method}

\vspace{-2mm}
\section{Dataset Preparation and Collection}
\vspace{-2mm}
\input{sec4_dataset}

\section{Experiments}
\input{sec5_exp}

\section{Conclusion}
\label{sec:conclusion}
In this paper, we introduced APIGen, a novel framework that generates reliable and diverse function-calling datasets by leveraging a multi-stage verification process. Our experiments demonstrate the effectiveness of APIGen in producing high-quality datasets from a wide range of executable APIs. This has significant implications for the development of more efficient and accessible language models, as it shows that high-quality data can be as important as model size in achieving strong performance.
By enabling smaller models to achieve competitive results and significantly enhancing their function-calling capabilities, our approach and released dataset open up new possibilities for the development of efficient and powerful language models in the agent tool-use domains. 

However, the current version of APIGen and the generated dataset have some limitations. Presently, the framework and dataset only consider REST APIs and Python functions. Additionally, although APIGen is a general framework, it currently only implements the generation procedure for single-turn function calling.
Future work will focus on extending APIGen to support more scenarios, programming languages, and APIs. We also plan to extend the framework to handle multi-turn and more complex interactions between agents and tools. Despite these limitations, we believe that APIGen and the generated dataset represent a significant step forward in the development of efficient and effective function-calling agents.

\bibliographystyle{unsrt}
\bibliography{main}

\input{checklist}
\clearpage
\appendix
\input{app-1}
\input{app-2}
\end{document}

%% file: sec1_intro_new.tex
\begin{figure}
  \centering
\includegraphics[width=0.98\textwidth]{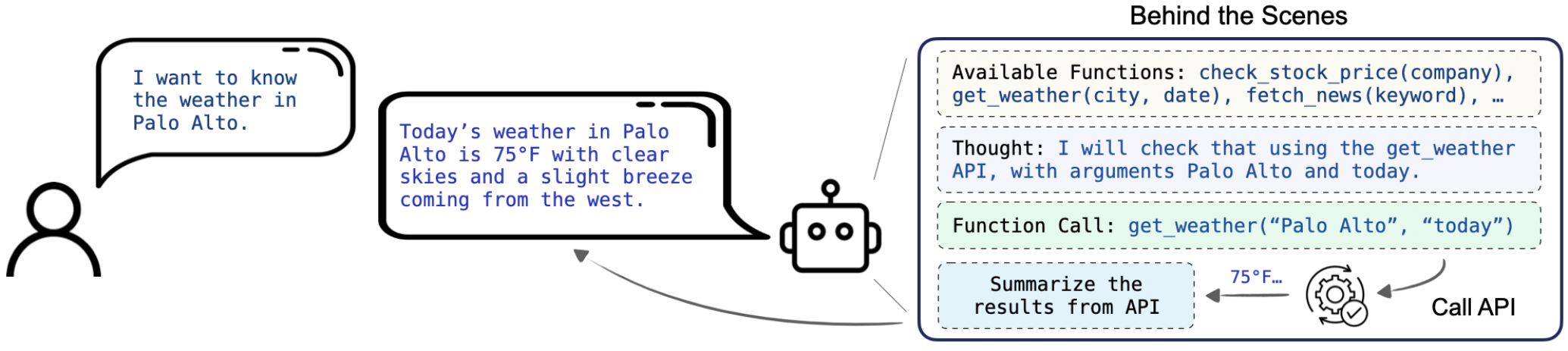}
  \caption{Workflow of an LLM-based function-calling agent.}
  \label{fig:workflow}
\end{figure}

Function-calling agents represent a significant advancement in artificial intelligence, specifically within the realm of Large Language Models (LLMs). These models, such as GPT4 \cite{achiam2023gpt}, Gemini \cite{reid2024gemini}, and Mistral \cite{jiang2024mixtral}, have evolved to not only understand and generate human-like text but also to execute functional API calls based on natural language instructions. For instance, consider a user requesting the weather in Palo Alto, as illustrated in Fig. \ref{fig:workflow}. The function-calling agent interprets this query, accesses the relevant API—such as \texttt{get\_weather("Palo Alto", "today")}—and retrieves the weather information, all in real-time. This capability extends the utility of LLMs beyond simple conversation tasks to include dynamic interactions with a variety of digital services and applications, ranging from social media platforms to financial services \cite{patil2023gorilla, srinivasan2023nexusraven, liu2023bolaa, chen2023agentverse, zhou2023webarena}.

Despite their growing popularity and potential, the deployment of function-calling agents is often hampered by the quality of the datasets used for training. Current datasets are largely static and lack comprehensive verification, leading to potential inaccuracies and inefficiencies of model fine-tuning in real-world applications \cite{bfcl, qin2023toolllm, wang2023mint, li2023api}. This limitation is particularly evident when models trained on these datasets encounter new, unseen APIs.
For example, a model trained primarily on restaurant booking APIs may struggle when suddenly tasked with retrieving stock market data, as it lacks the specific training data or the adaptability to handle new domains.

To address these challenges, we introduce APIGen, an \textbf{A}utomated \textbf{PI}peline for \textbf{Gen}erating verifiable and diverse function-calling datasets. Our framework is designed to facilitate the fine-tuning of function-calling LLMs by providing high-quality, diverse datasets that better reflect the variability and complexity of real-world API use. 
Crucially, each generated data point undergoes rigorous multi-stage verification processes—format, execution, and semantic—to ensure accuracy and applicability.
We fine-tune function-calling models using the dataset generated by APIGen. The results show the strong performance of our models, surpassing many existing powerful LLMs with much fewer parameters, highlighting the effectiveness of APIGen and the high quality of the dataset it produces.

With APIGen, we release a comprehensive dataset containing 60,000 entries with 3,673 APIs across 21 categories. They include various query styles, such as parallel function calling data (asking the agent to produce multiple concurrent function calls in a single response) \cite{bfcl}, which is rarely found in public datasets, to the best of our knowledge. This large-scale synthetic dataset is intended to catalyze further research and development in the field of function-calling agents, offering researchers and developers a foundation for training and testing their models. The data is available on Huggingface and our project homepage.

The contributions of this work are summarized as follows:
\begin{itemize}[leftmargin=10pt]
    \item We introduce APIGen, a function-calling data generation pipeline that features verifiability, scalability, and diversity of the data. APIGen is compatible with a range of models and APIs to construct high-quality synthetic function-calling datasets.
    \item We train two function-calling models of different sizes, 1.3B and 6.7B, using APIGen-constructed training data. Extensive experiments demonstrate that the 6.7B model achieves a rank of 6th on the Berkeley Function-Calling Leaderboard \cite{bfcl}, surpassing GPT-4o and Gemini-1.5-Pro, while the 1.3B model outperforms GPT-3.5-Turbo.
    \item We also release a synthetic function-calling dataset containing 60,000 high-quality data generated by APIGen using several strong open-source LLMs, which can potentially benefit the research community in developing advanced function-calling models.
\end{itemize}

%% file: sec2_related_new.tex
\textbf{Tool-use Agent.} 
Recent works have developed frameworks and models that 
enable LLMs to interact with APIs and tools \cite{wang2023voyager,hong2023metagpt,chen2023fireact,xu2023creative,xu2023lemur,zeng2023agenttuning,liu2023tail,he2024webvoyager}. RestGPT \cite{song2023restgpt} connects LLMs to RESTful APIs using a Planner, API selector, and API executor to handle complex instructions. Toolformer \cite{schick2024toolformer} is an early work that enables agents to use tools like Question Answering, Calculator, and Wikipedia Search through a supervised-finetuned model. 
\cite{zhang2024agentohana} designs the xLAM model, showing strong tool usage capability across several benchmarks. Octopus-v4 \cite{chen2024octopus} presents a methodology to incorporate multiple specialized language models to solve corresponding tasks. While 
NexusRaven \cite{srinivasan2023nexusraven} and Gorilla OpenFunctions-v2 \cite{gorilla-openfunctions-v2} are strong open-sourced models that focus on function calling, neither provides access to their training datasets.

\textbf{Agent Datasets.}
Several datasets have been created to support the development of agent models. AgentInstruct \cite{zeng2023agenttuning} consists of 6 datasets for different agent tasks, including AlfWorld \cite{shridhar2020alfworld}, WebShop \cite{yao2022webshop}, Mind2Web \cite{deng2024mind2web}, Knowledge Graph, Operating System, and Database \cite{liu2023agentbench}. APIBank \cite{li2023api} is a benchmark designed for tool-augmented LLMs, providing a training set containing tool-use dialogues from various APIs. Toolalpaca \cite{tang2023toolalpaca} constructs a varied and well-structured tool-use dataset by randomly selecting APIs and generating documentation using ChatGPT. ToolBench \cite{qin2023toolllm} creates an instruction-tuning dataset for tool use by collecting diverse REST APIs and generating their descriptions using ChatGPT. AgentOhana \cite{zhang2024agentohana} and Lumos \cite{yin2023lumos} design a unified data and training pipeline for efficient agent learning, covering multiple different datasets and environments. However, most of these datasets were not rigorously verified, and usually contain noisy data.

\textbf{Benchmarks.}
Recent studies have established several benchmarks to assess agent abilities on various tasks such as web interactions, reasoning, decision making, function calling, and tool usage \citep{yao2022webshop,zhou2023webarena, liu2023agentbench,chen2023agentverse,huang2023metatool,qin2023toolllm,wang2023mint,liu2023datasets,ma2024agentboard,liu2024agentlite}. Specifically, AgentBoard \cite{ma2024agentboard} includes 9 tasks, with ToolOperation and ToolQuery designed to evaluate agent ability on multi-turn interaction with external tools. ToolEval \cite{qin2023toolllm} assesses functional calling capabilities via RapidAPI, containing around 1,000 test cases and asking GPT-3.5 to assess the Win Rate. Furthermore, the Berkeley Function-Calling Leaderboard (BFCL) \cite{bfcl} provides a robust and comprehensive framework to evaluate models' abilities to call functions, with 2,000 test cases covering a wide range of scenarios. We use BFCL as our testing ground as it provides the most thorough comparison among popular LLMs.

%% file: sec3_method.tex
\begin{figure}
  \begin{center}
  \includegraphics[width=0.99\linewidth]{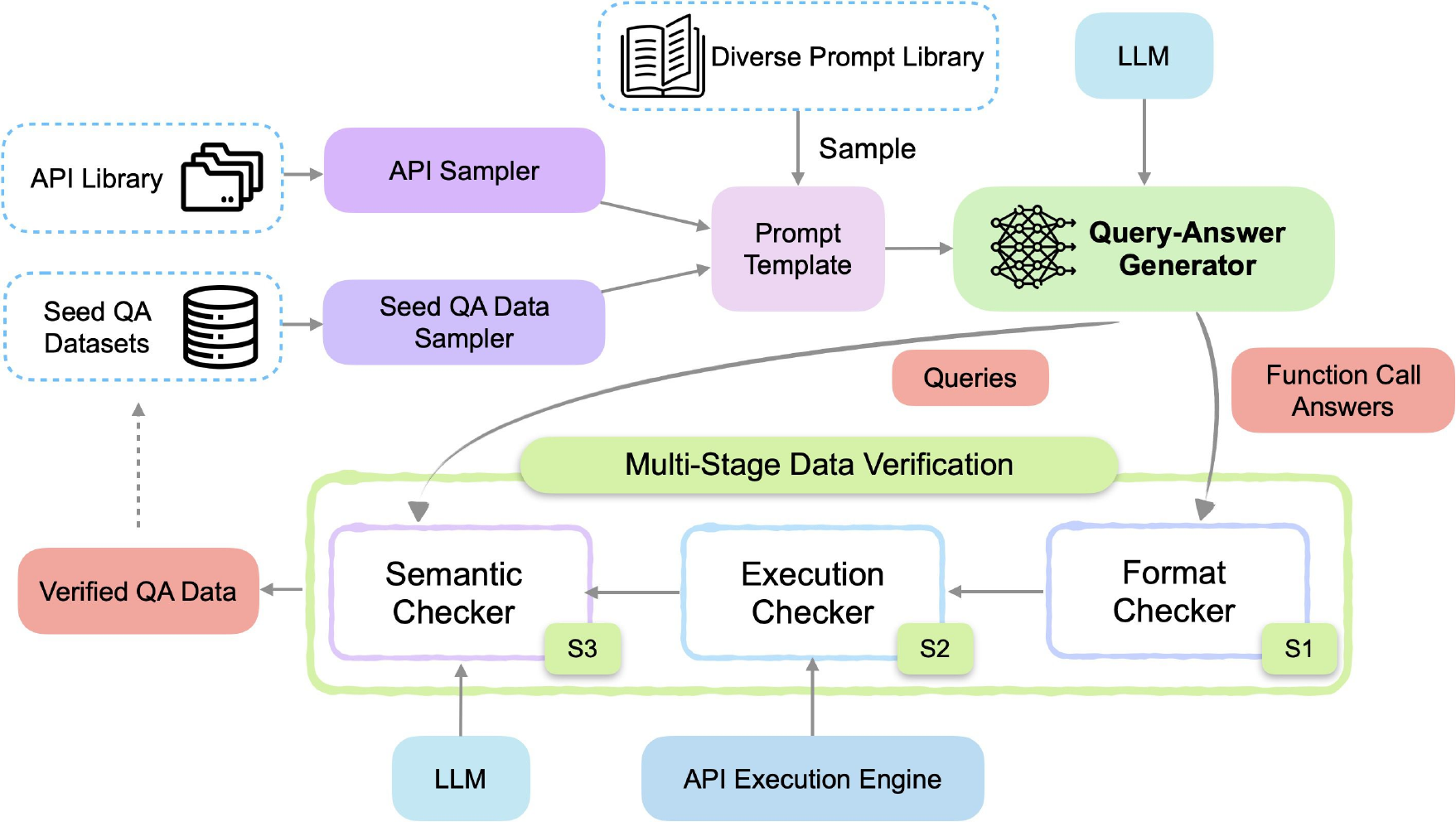}
  \end{center}
  \caption{\small Illustration of the post-process filters.}
  \label{fig:overview}
\end{figure}

This section introduces the detailed design of APIGen, an \textbf{A}utomated \textbf{PI}peline for \textbf{Gen}erating verifiable and diverse function-calling datasets. Our framework is designed with three key factors in mind: data quality, data diversity, and collection scalability. We achieve these through the key modules shown in Fig.~\ref{fig:overview}: the multi-stage data verification process ensures data quality, the seed QA (query-answer) data sampler, API sampler, and various prompt templates ensure diversity, and our structured modular design using a unified format enables the system to scale to diverse API sources, including but not limited to Python functions and representational state transfer (REST) APIs.

\subsection{Data Generation Overview}

\begin{figure}
  \begin{center}
\includegraphics[width=0.99\linewidth]{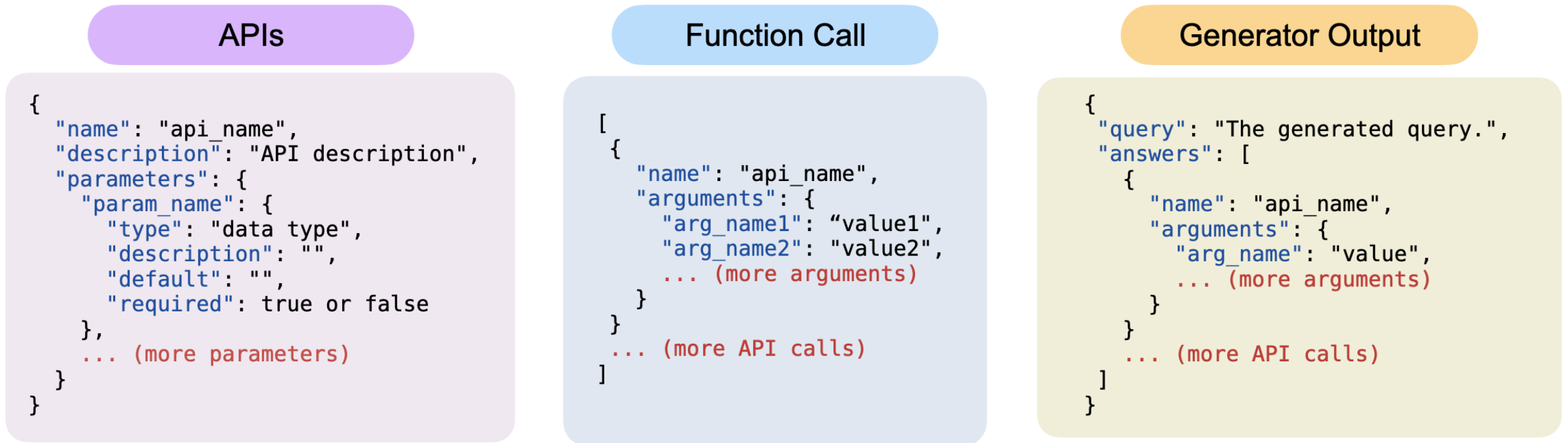}
  \end{center}
  \caption{\small JSON data format examples.}
  \label{fig:json_format}
\end{figure}

Figure \ref{fig:overview} outlines the data generation process using the APIGen framework, which begins by sampling one or more APIs and example query-answer (QA) pairs (seed data) from the library, then formatting them into a standardized JSON format (see Fig. \ref{fig:json_format} for examples). A prompt template is selected based on the desired data generation objectives, which steers the LLM in generating relevant query-answer pairs. Each answer in the generated pairs is a function call formatted in JSON.

The adoption of a standardized JSON format for APIs, function calls, and generator outputs (as shown in Figure \ref{fig:json_format}) provides several advantages. Firstly, it establishes a structural way to verify whether the generator's output contains all necessary fields. Outputs that fail to comply with these format requirements are discarded. Secondly, the JSON structure enables efficient checking of function calls for correct parsing and validity of arguments. Calls that include arguments not present in the API library or hallucinate non-existent functions are excluded, enhancing the overall quality of the dataset.
Another key benefit is the scalability it enables. With this uniform format, APIGen can easily incorporate data from diverse sources (Python functions, REST APIs, etc) by developing format converters that adapt them into these basic JSON elements, without modifying other core components, such as the prompting library, making the framework highly adaptable and extensible.

The generated function calls are subjected to a multi-stage verification process to ensure their correctness and relevance.
First, a format checker verifies correct JSON formatting and parseability. Next, the API execution engine processes the calls and sends the results and queries to a semantic checker, another LLM, which assesses alignment between the function calls, execution results, and query objectives. Data points passing all stages are added back to the seed dataset as high-quality examples to enhance future generation diversity.
We detail each checker in the next section.

\subsection{Multi-Stage Data Verification}

Prioritizing quality is crucial, as previous research has shown that small amounts of high-quality fine-tuning data can substantially enhance model performance on domain-specific tasks \cite{zhou2024lima}. This motivates our multi-stage dataset verification process to align large language models effectively.

The key insight driving our framework design is that, unlike synthetic chat data which can be difficult to evaluate, function-calling answers can be directly executed via their corresponding APIs. This enables checking if the output API and parameters' formats are correct, if the generated API calls are executable, and if execution results match the query's intent, etc. Based on this observation, we propose a three-stage verification process:

\textbf{Stage 1: Format Checker}: This stage performs sanity checks to filter out poorly formatted or incomplete data. The LLM output must strictly follow a JSON format with the "query" and "answer" fields, as shown in Fig. \ref{fig:json_format}. We usually also include an additional "thought" field, which is known as the chain-of-thought (CoT) prompting technique \cite{wei2022chain}, to increase the pass rate of the generated data. The data is discarded if these fields cannot be properly extracted for function calls. Additionally, the function calls are checked for correct JSON parsing and valid arguments. Generated calls whose arguments or functions are not present in the given APIs are eliminated to reduce hallucination and improve data quality.

\textbf{Stage 2: Execution Checker}: Well-formatted function calls from Stage 1 are executed against the appropriate backend (e.g. Python functions are directly imported and executed in a separate subprocess, while REST APIs are called to obtain results and status codes). Unsuccessful executions are filtered out, and fine-grained error messages are provided for failures, including argument type errors, invalid parameters, runtime errors, timeout, syntax errors, missing arguments, etc.

\textbf{Stage 3: Semantic Checker}: Successful Stage 2 execution results, available functions, and the generated query are formatted and passed to another LLM to assess if the results semantically align with the query's objective. Query-answer pairs that execute successfully but produce meaningless results due to infeasible queries or incorrect arguments are filtered out. The main decision factors for this stage are: 1) whether the function call aligns with the query's objective and has proper arguments; 2) whether the function call and arguments are appropriately chosen from the available functions; 3) whether the number of function calls matches the user's intent; 4) whether the execution results contain errors or indicate unsuccessful function execution; 5)  whether the execution results are relevant and match the query's purpose. APIGen's design offers the flexibility to select one or multiple LLMs as checkers, and the filtering rules can be readily adjusted—either tightened or relaxed—depending on specific use cases.

Data points that pass all three verification stages are regarded as high-quality and added back to improve future diverse data generation. This multi-stage verification process is the key to ensuring the APIGen framework produces a dataset that is not only diverse but also of verifiable high quality, enabling more effective fine-tuning of LLMs to domain-specific API-related tasks.

\subsection{Methods to Improve Dataset Diversity}
\label{sec:diversity}

Encouraging diversity in training datasets is crucial for developing robust function-calling agents that can handle a wide range of real-world scenarios. In APIGen, we promote data diversity through multiple perspectives, including query style diversity, sampling diversity, and API diversity.

\textbf{Query Style Diversity.} 
APIGen's dataset is structured into four main categories: simple, multiple, parallel, and parallel multiple, each designed to challenge and enhance the model's capabilities in different usage scenarios. These categories are inspired by the Berkeley function-calling benchmark \cite{bfcl} and are controlled by corresponding prompts and seed data. We show examples of them in the supplementary material. The categories are as follows:

\begin{itemize}[leftmargin=10pt]
\item \textbf{Simple:} This query style includes straightforward scenarios where a single function call is made based on the user's input with a single provided JSON format API description. 
\item \textbf{Multiple:} In this style, user queries could be answered by one of several function calls. The challenge lies in selecting the most appropriate function from multiple provided APIs. It represents one of the most common real-world use cases.
\item \textbf{Parallel:} This query style requires executing multiple function calls simultaneously in response to a single user query, which may consist of one or more sentences but with only one API provided. For instance, if the user wants to know the weather in both \texttt{Palo Alto} and \texttt{Paris}, the model should call the \texttt{get\_weather} function twice with corresponding city names in a single response.
\item \textbf{Parallel Multiple:} This query style combines the parallel and multiple categories, where multiple function and API documents are provided, and each function call might be invoked multiple times based on the query's requirements.
\end{itemize}

While there exist publicly available training data for \textit{simple} and \textit{multiple} categories \cite{guo2024stabletoolbench, qin2023toolllm}, however,
to the best of our knowledge, we offer the first large-scale and high-quality datasets that include the \textit{parallel}-related function-calling scenario.

\textbf{Sampling Diversity.}
APIGen utilizes a sampling system designed to maximize the diversity and relevance of the generated datasets, which include three main components, as shown in Fig. \ref{fig:overview}:

\begin{itemize}[leftmargin=10pt]
    \item \textbf{API Sampler:} This module extracts one or more function descriptions from executable API libraries, standardizing them into a uniform JSON format. The diverse sources of APIs ensure a wide range of function calls are available for inclusion in the training dataset.
    \item \textbf{Example Sampler:} It samples a specified number of seed examples corresponding to the different categories. These examples are transformed into structured queries, function descriptions, and answers, serving as an important few-shot reference for data generation.
    \item \textbf{Prompt Sampler:} This sampler draws from a diverse prompt library to generate a variety of query-answer pairs. 
    The prompts for each query style contain different contexts, ranging from simple, concise query-answer pairs to more realistic scenarios, such as ambiguous or misspelled user requests, enhancing the model's ability to handle real-world interactions.
\end{itemize}

We provide some prompt templates and seed data in the supplementary material.
In APIGen, the number of examples and APIs sampled for each dataset iteration is randomly chosen from a predefined range. This randomization enhances dataset variability by preventing repetitive patterns and ensuring a broad coverage of scenarios. We next introduce our API diversity.

%% file: sec4_dataset.tex
We begin by discussing our dataset preparation process, which includes selecting and cleaning API libraries. Then we present our dataset collection setup and an overview of the resulting dataset.

\subsection{Dataset API Sources}

To ensure a high-quality and diverse dataset, we focused on collecting real-world APIs that could be readily executed and came with thorough documentation. We primarily sourced APIs from ToolBench \cite{qin2023toolllm}, a comprehensive tool-use dataset that includes 16,464 REST APIs across 49 coarse-grained categories from RapidAPI Hub. This hub is a leading marketplace featuring a vast array of developer-contributed APIs. 
To further enhance the usability and quality of the APIs, we perform the following filtering and cleaning procedures on the ToolBench dataset:
\begin{itemize}[leftmargin=10pt]
\item \textbf{Data Quality Filtering:} We remove APIs with incorrectly parsed documentation and those lacking required or optional parameters. APIs requiring no parameters were excluded to maintain the challenge level appropriate for our dataset needs.
\item \textbf{API Accessibility Testing:} We tested API accessibility by making requests to each endpoint using example parameters provided in the dataset and through the Stable Toolbench server \cite{guo2024stabletoolbench}. APIs that could not be executed or returned errors, such as timeouts or invalid endpoints, were discarded.
\item \textbf{Docstring Regeneration:} To improve the quality of API documentation, we regenerated docstrings for the APIs that have noisy and unusable descriptions.
\end{itemize}

\begin{wrapfigure}{r}{0.5\linewidth}
  \begin{center}
  \includegraphics[width=\linewidth]{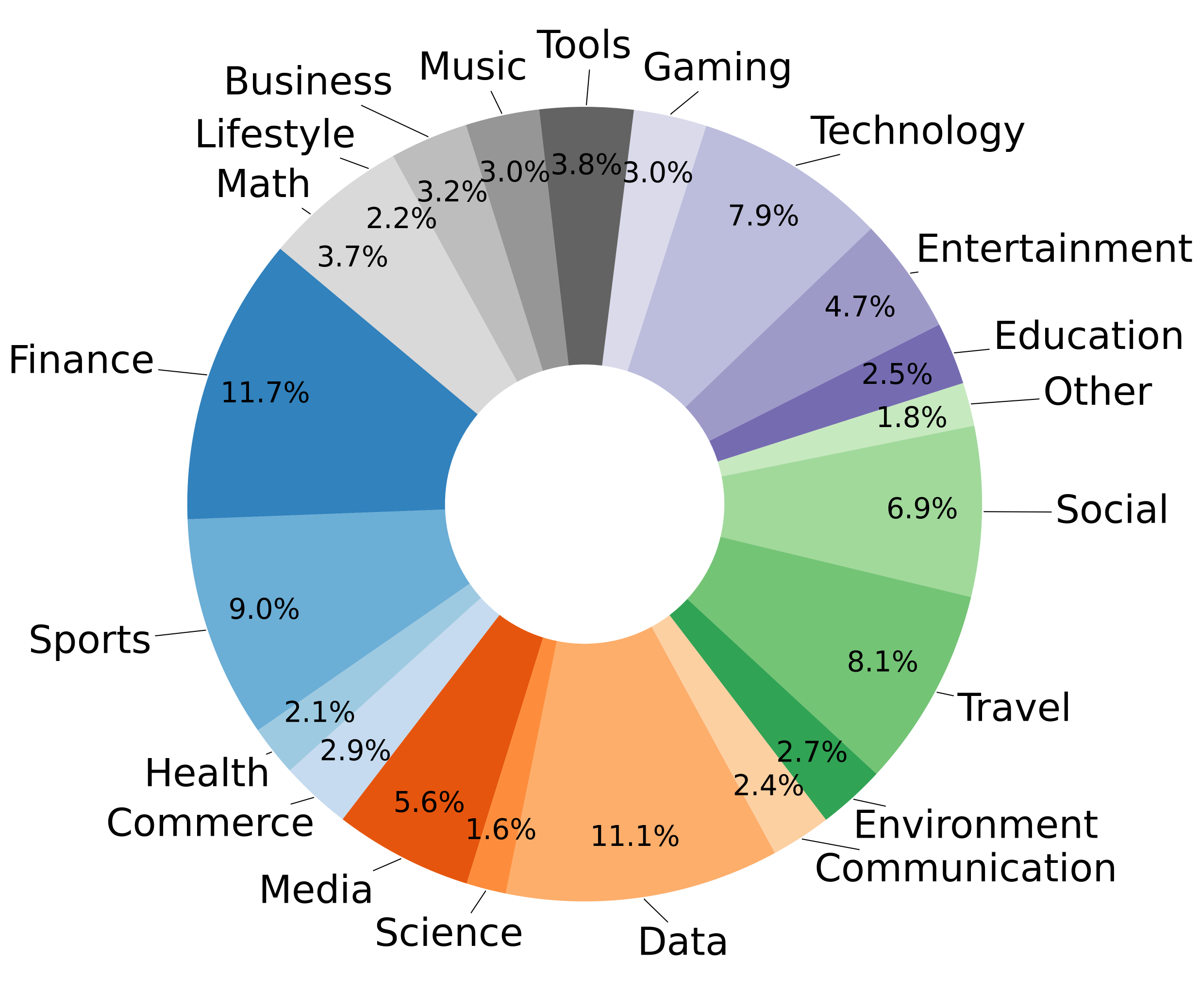}
  \end{center}
  \caption{\small The category distribution of the 3,673 executable APIs.}
  \label{fig:dataset_category}
\end{wrapfigure}

After cleaning, we obtain 3,539 executable REST APIs with good documentation. 
Additionally, we incorporated Python functions as another API type, inspired by the executable evaluation categories of the Berkeley function-calling benchmark \cite{bfcl}. We collected 134 well-documented Python functions covering diverse fields such as mathematics, finance, and data management. Sample API examples are provided in the supplementary material.

The original ToolBench dataset contained semantically overlapping categories such as \texttt{Finance} and \texttt{Financial}. We consolidated these into 21 distinct categories to ensure clarity and balance across the dataset. Figure \ref{fig:dataset_category} illustrates the distribution of the 3,673 executable APIs across these redefined categories, spanning sectors like technology, social sciences, education, and sports.
This diverse collection of APIs provides a strong foundation for synthetic data generation and is a valuable asset for ensuring data quality and reliability.

\subsection{Collection Setup and Dataset Details}

To validate the effectiveness of the APIGen framework, we generated datasets targeting various query styles as outlined in Section \ref{sec:diversity}. We utilized several base LLMs for data generation, including DeepSeek-V2-Chat (236B) \cite{deepseekv2}, DeepSeek-Coder-33B-Inst \cite{guo2024deepseek}, Mixtral-8x22B-Inst, and Mixtral-8x7B-Inst \cite{jiang2024mixtral}. For each model, our target was to generate 40,000 data points by sampling different combinations of APIs, seed data, and prompt templates. To foster diversity in the generated responses, we set the generation temperature to 0.7 across all models. Examples of the prompt templates and APIs used are provided in the supplementary materials for reference.

Table \ref{tab:filtering} presents statistics for the data generation process with different models, including the total verified data point count and the number of filtered data points at each verification stage. The filtering process successfully removes many low-quality data points due to formatting issues, execution errors, or failure to pass the semantic check. The first two stages, format checker and execution checker, typically filter out the majority of low-quality data. These data points often have infeasible argument ranges, incorrect types, missing required parameters, or more severe issues such as hallucination of function calls or parameters. Our systematic verification process provides a rigorous way to reduce the occurrence of these situations.

\begin{table}[ht]
\centering
\caption{Filtering statistics for the generated datasets using different base LLMs.}
\resizebox{1.\linewidth}{!}{
\begin{tabular}{@{}c|c|ccc|c@{}}
\toprule
Model & Verified Data & Fail Format & Fail Execution & Fail Semantic & Pass Rate \\ \midrule
DeepSeek-Coder-33B-Inst & 13,769 & 4,311 & 15,496 & 6,424 & 34.42\% \\
Mixtral-8x7B-Inst & 15,385 & 3,311 & 12,341 & 7,963 & 38.46\% \\
Mixtral-8x22B-Inst & 26,384 & 1,680 & 5,073 & 6,863 & 65.96\% \\
DeepSeek-V2-Chat (236B) & 33,659 & 817 & 3,359 & 2,165 & 84.15\% \\\bottomrule
\end{tabular}
}
\label{tab:filtering}
\end{table}

The semantic checker also plays a crucial role in filtering generated data that does not align with the query's objectives. For instance, if a user's query contains multiple requests, but the returned results only address one, or if the generated function-call data and execution results do not match the user's query, the data point will be filtered out. Including these data points in the training set for model training could potentially harm the performance, as demonstrated in the experiments.

We observe that stronger models like DeepSeek-V2-Chat and Mixtral-8x22B-Inst have better format-following capabilities and higher pass rates, while the two relatively smaller models have a much higher likelihood of producing data that cannot be executed. This suggests that when using weaker models to generate data, a strict verification process is recommended to filter out low-quality data.

We are releasing approximately 60,000 high-quality function-calling datasets generated from the two strongest models: Mixtral-8x22B-Inst and DeepSeek-V2-Chat (236B). These datasets include all the query styles mentioned in Sec. \ref{sec:diversity} and cover a wide range of practical situations, with 3,673 diverse APIs across 21 categories. Each data point has been verified using real-world APIs to ensure its validity and usefulness. By making this dataset publicly available, we aim to benefit the research community and facilitate future work in this area.

%% file: sec5_exp.tex
\subsection{Experiment Setup}

To evaluate the utility and effectiveness of the collected dataset, we conducted experiments by training function-calling models with the generated data. Our aim is to answer two key questions:
1) To what extent can the generated data boost the model's function-calling capability, and how does it compare to existing models?
2) How effective is the APIGen framework in filtering out low-quality data?

To address these questions, we train two versions of base models: DeepSeek-Coder-1.3B-instruct and DeepSeek-Coder-7B-instruct-v1.5 \cite{guo2024deepseek} using the xLAM (large action model) training pipeline proposed in \citep{zhang2024agentohana}. We refer to these models as xLAM-1B (FC) and xLAM-7B (FC), where FC stands for the Function-Calling mode, similar to this mode in other existing models that output JSON-format function calls \cite{achiam2023gpt,gorilla-openfunctions-v2,cohere_command_r_plus,patil2023gorilla}. We compare the performance of these small-sized models against state-of-the-art models, including different versions of GPT-4 series \cite{achiam2023gpt}, Claude-3 series \cite{anthropic2024claude}, Gemini series \cite{reid2024gemini}, Llama3 \cite{touvron2023llama}, Mixtral \cite{jiang2024mixtral}, OpenFunctions-v2 \cite{gorilla-openfunctions-v2}, Command R+ \cite{cohere_command_r_plus}, etc.

\textbf{Benchmark.} We evaluate the trained models' performance on the Berkeley Function-Calling Benchmark (BFCL) \cite{bfcl}, which provides a comprehensive evaluation framework for assessing the function-calling capabilities of LLMs across various programming languages and application domains. Designed to reflect real-world use cases, the BFCL includes 2,000 testing cases, covering complex scenarios such as parallel and multiple-function calls. The benchmark contains diverse API sources like Java, JavaScript, and Python, offering a detailed analysis of each model's ability to correctly interpret and execute commands under different conditions.
BFCL serves as a highly detailed and scalable benchmark for evaluating LLMs' function-calling capabilities and provides a leaderboard to track the most recent and powerful LLMs, both commercialized and open-source.

\textbf{Evaluation Metrics.}
The Berkeley Function-Calling Leaderboard (BFCL) evaluates LLMs using two main categories: Abstract Syntax Tree (AST) Evaluation and Executable Function Evaluation. The AST evaluation focuses on the syntactic accuracy of the generated function calls, ensuring that the model's output matches a predefined function documentation in structure and parameters. This includes checks for correct function names, required parameters, and appropriate data types. The Executable Function Evaluation goes a step further by running the generated function calls to verify their operational correctness. This executable test ensures that the functions not only compile but also execute correctly, providing the expected results, which is crucial for practical applications where real-time performance is essential.

\subsection{Experiment Results Analysis}

\textbf{Can the generated data improve the model's function-calling capability and how does it compare to other most powerful models?}
The performance of our models, xLAM-7B and xLAM-1B, as presented in Table \ref{table:comparison}, highlights the effectiveness of our APIGen framework and the quality of the datasets produced. 
Notably, our xLAM-7B model ranks 6th among the most powerful LLMs listed on the BFCL leaderboard~\footnote{\url{https://gorilla.cs.berkeley.edu/leaderboard.html\#leaderboard}}, surpassing several versions of GPT-4 (GPT-4o, GPT4-Turbo-FC), Llama3-70B, multiple Claude-3 models, and a series of strong models which are known for their exceptional capabilities in various tasks, including function-calling.
This achievement demonstrates the significant impact of our high-quality dataset on the model's function-calling performance.

\input{tables}

Our smaller xLAM-1B model also shows remarkable results, securing the 24th position and outperforming many larger models, such as Claude-3 Haiku \cite{anthropic2024claude}, Command-R-Plus \cite{cohere_command_r_plus}, DBRX-Instruct \cite{databricks_dbrx_instruct}, Mistral-large \cite{jiang2024mixtral}, and GPT-3.5-Turbo-0125.
The results highlight the effectiveness of the APIGen pipeline in enhancing a model's function-calling capabilities, even with a much smaller size. 
Both xLAM-7B and xLAM-1B demonstrate substantial improvements in handling complex query types, particularly in the \textit{parallel} and \textit{multiple} function-calling scenarios, which are typically underrepresented in existing publicly available dataset. This validates the value of our pipeline and datasets in addressing practical scenarios involving complex API interactions and multiple concurrent API calls, especially considering that the base model, DeepSeek-Coder-v1.5, only ranks 45th on the leaderboard and performs poorly in these categories.

\begin{wrapfigure}{r}{0.5\linewidth}
  \begin{center}
  \vspace{-6mm}
  \includegraphics[width=0.99\linewidth]{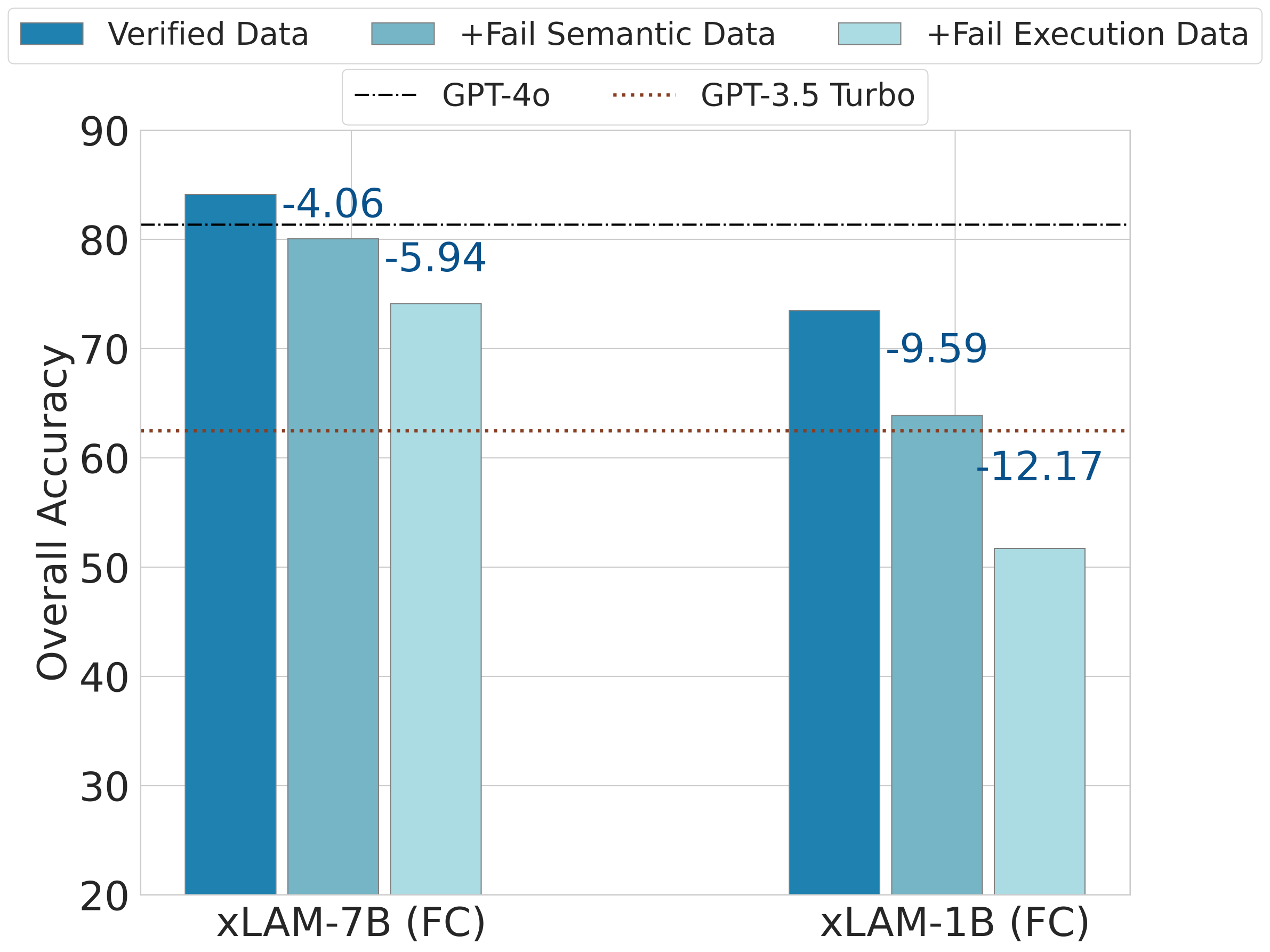}
  \end{center}
  \vspace{-2mm}
  \caption{\small Performance comparison of using different stage's datasets from APIGen. ``+Fail Semantic Data" and ``+Fail Execution Data" meaning adding the filtered dataset from stage 3 and stage 2 to the training set.}
  \label{fig:ablation}
  \vspace{-6mm}
\end{wrapfigure}

Next, we answer the question: \textbf{how effective is the APIGen framework in filtering out low-quality data?} We conducted an ablation study by adding the datasets that were filtered out by stage 3 (semantic checker) and stage 2 (execution checker) back to the training set, simulating situations where generated data is used without the rigorous verification process. The performance comparison on the BFCL benchmark, shown in Fig. \ref{fig:ablation}, reveals that using these filtered datasets for training harms the final performance, with a more significant impact on the smaller model. This indicates that directly using generated data might not yield the best results and demonstrates the effectiveness of our APIGen framework in filtering out low-quality data.

These results provide compelling evidence for the effectiveness of the APIGen framework in generating high-quality, diverse datasets for function-calling tasks. The impressive performance achieved by our small-sized models highlights the efficiency of our approach, demonstrating that by focusing on data quality and diversity, we can effectively boost the performance of smaller models, making them competitive with much larger ones in this function-calling agent domain.

%% file: tables.tex
\begin{table}[h]
\centering
\caption{\small Performance comparison of different models on BFCL leaderboard (as of date 06/15/2024). The rank is based on the overall accuracy, which is a weighted average of different evaluation categories. ``FC" stands for function-calling mode in contrast to using a customized ``prompt" to extract the function calls. See the benchmark \cite{bfcl} for details.}
\label{table:comparison}
\resizebox{1.\linewidth}{!}{
\begin{tabular}{|cccccccccccc|}
\hline
\multicolumn{1}{|c|}{}                           & \multicolumn{1}{c|}{}                                                                             & \multicolumn{1}{c|}{}                                     & \multicolumn{4}{c|}{}                                                                                                                                                                       & \multicolumn{4}{c|}{}                                                                                                                                                                       &                                                                                 \\
\multicolumn{1}{|c|}{}                           & \multicolumn{1}{c|}{}                                                                             & \multicolumn{1}{c|}{}                                     & \multicolumn{4}{c|}{\multirow{-2}{*}{Abstract Syntax Tree (AST) Evaluation}}                                                                                                                & \multicolumn{4}{c|}{\multirow{-2}{*}{Evaluation by Executing APIs}}                                                                                                                         &                                                                                 \\ \cline{4-11}
\multicolumn{1}{|c|}{}                           & \multicolumn{1}{c|}{}                                                                             & \multicolumn{1}{c|}{}                                     &                              &                            &                            & \multicolumn{1}{c|}{}                                                                              &                              &                            &                            & \multicolumn{1}{c|}{}                                                                              &                                                                                 \\
\multicolumn{1}{|c|}{\multirow{-4}{*}{Rank}}     & \multicolumn{1}{c|}{\multirow{-4}{*}{\begin{tabular}[c]{@{}c@{}}Overall\\ Accuracy\end{tabular}}} & \multicolumn{1}{c|}{\multirow{-4}{*}{Model}}              & \multirow{-2}{*}{Simple}     & \multirow{-2}{*}{Multiple} & \multirow{-2}{*}{Parallel} & \multicolumn{1}{c|}{\multirow{-2}{*}{\begin{tabular}[c]{@{}c@{}}Parallel\\ Multiple\end{tabular}}} & \multirow{-2}{*}{Simple}     & \multirow{-2}{*}{Multiple} & \multirow{-2}{*}{Parallel} & \multicolumn{1}{c|}{\multirow{-2}{*}{\begin{tabular}[c]{@{}c@{}}Parallel\\ Multiple\end{tabular}}} & \multirow{-4}{*}{\begin{tabular}[c]{@{}c@{}}Relevance\\ Detection\end{tabular}} \\ \hline
\multicolumn{1}{|c|}{1}                          & \multicolumn{1}{c|}{88}                                                                           & \multicolumn{1}{c|}{GPT-4-0125-Preview (Prompt)}          & 88.36                        & 95                         & 90.5                       & \multicolumn{1}{c|}{91}                                                                            & 99.41                        & 94                         & 84                         & \multicolumn{1}{c|}{75}                                                                            & 70.42                                                                           \\
\multicolumn{1}{|c|}{2}                          & \multicolumn{1}{c|}{87.65}                                                                        & \multicolumn{1}{c|}{Claude-3-Opus-0229 (Prompt)}          & 86.73                        & 94                         & 86                         & \multicolumn{1}{c|}{89}                                                                            & 97.65                        & 92                         & 80                         & \multicolumn{1}{c|}{75}                                                                            & 80.42                                                                           \\
\multicolumn{1}{|c|}{3}                          & \multicolumn{1}{c|}{86.35}                                                                        & \multicolumn{1}{c|}{Gemini-1.5-Pro-0514 (FC)}             & 80.18                        & 92                         & 91                         & \multicolumn{1}{c|}{88.5}                                                                          & 91.76                        & 88                         & 76                         & \multicolumn{1}{c|}{77.5}                                                                          & 89.58                                                                           \\
\multicolumn{1}{|c|}{4}                          & \multicolumn{1}{c|}{85.88}                                                                        & \multicolumn{1}{c|}{Gemini-1.5-Pro-0409 (FC)}             & 80                           & 92.5                       & 90                         & \multicolumn{1}{c|}{88}                                                                            & 90                           & 90                         & 74                         & \multicolumn{1}{c|}{77.5}                                                                          & 88.75                                                                           \\
\multicolumn{1}{|c|}{5}                          & \multicolumn{1}{c|}{85.88}                                                                        & \multicolumn{1}{c|}{GPT-4-1106-Preview (FC)}              & 84                           & 91.5                       & 92                         & \multicolumn{1}{c|}{87}                                                                            & 89.41                        & 92                         & 78                         & \multicolumn{1}{c|}{67.5}                                                                          & 80.42                                                                           \\
\rowcolor[HTML]{C0C0C0} 
\multicolumn{1}{|c|}{\cellcolor[HTML]{C0C0C0}6}  & \multicolumn{1}{c|}{\cellcolor[HTML]{C0C0C0}85.65}                                                & \multicolumn{1}{c|}{\cellcolor[HTML]{C0C0C0}xLAM-7B (FC)} & {\color[HTML]{1F1F1F} 80.55} & {\color[HTML]{1F1F1F} 96}  & {\color[HTML]{1F1F1F} 90}  & \multicolumn{1}{c|}{\cellcolor[HTML]{C0C0C0}{\color[HTML]{1F1F1F} 87.5}}                           & {\color[HTML]{1F1F1F} 90.59} & {\color[HTML]{1F1F1F} 90}  & {\color[HTML]{1F1F1F} 86}  & \multicolumn{1}{c|}{\cellcolor[HTML]{C0C0C0}{\color[HTML]{1F1F1F} 77.5}}                           & {\color[HTML]{1F1F1F} 80.42}                                                    \\
\multicolumn{1}{|c|}{7}                          & \multicolumn{1}{c|}{85.59}                                                                        & \multicolumn{1}{c|}{GPT-4-turbo-20240409 (Prompt)}        & 86.55                        & 95                         & 90                         & \multicolumn{1}{c|}{88.5}                                                                          & 97.65                        & 94                         & 80                         & \multicolumn{1}{c|}{72.5}                                                                          & 62.5                                                                            \\
\multicolumn{1}{|c|}{8}                          & \multicolumn{1}{c|}{84.71}                                                                        & \multicolumn{1}{c|}{Gorilla-OpenFunctions-v2 (FC)}        & 88                           & 95                         & 87.5                       & \multicolumn{1}{c|}{87}                                                                            & 94.71                        & 94                         & 70                         & \multicolumn{1}{c|}{67.5}                                                                          & 61.25                                                                           \\
\multicolumn{1}{|c|}{9}                          & \multicolumn{1}{c|}{84.65}                                                                        & \multicolumn{1}{c|}{GPT-4-0125-Preview (FC)}              & 80.18                        & 93                         & 90.5                       & \multicolumn{1}{c|}{85}                                                                            & 83.53                        & 92                         & 86                         & \multicolumn{1}{c|}{77.5}                                                                          & 82.92                                                                           \\
\multicolumn{1}{|c|}{10}                         & \multicolumn{1}{c|}{83.88}                                                                        & \multicolumn{1}{c|}{Llama-3-70B-Instruct (Prompt)}        & 81.45                        & 93                         & 91.5                       & \multicolumn{1}{c|}{85}                                                                            & 91.76                        & 88                         & 84                         & \multicolumn{1}{c|}{77.5}                                                                          & 69.17                                                                           \\
\multicolumn{1}{|c|}{11}                         & \multicolumn{1}{c|}{82.94}                                                                        & \multicolumn{1}{c|}{GPT-4o-2024-05-13 (FC)}               & 78.91                        & 90                         & 87.5                       & \multicolumn{1}{c|}{84.5}                                                                          & 86.47                        & 78                         & 82                         & \multicolumn{1}{c|}{75}                                                                            & 81.25                                                                           \\
\multicolumn{1}{|c|}{12}                         & \multicolumn{1}{c|}{82.88}                                                                        & \multicolumn{1}{c|}{GPT-4-turbo-2024-04-09 (FC)}          & 74.73                        & 90                         & 89.5                       & \multicolumn{1}{c|}{88}                                                                            & 82.94                        & 88                         & 76                         & \multicolumn{1}{c|}{67.5}                                                                          & 88.75                                                                           \\
\multicolumn{1}{|c|}{13}                         & \multicolumn{1}{c|}{81.82}                                                                        & \multicolumn{1}{c|}{Claude-3-Sonnet-20240229 (Prompt)}    & 83.09                        & 89                         & 88                         & \multicolumn{1}{c|}{89.5}                                                                          & 93.53                        & 92                         & 84                         & \multicolumn{1}{c|}{77.5}                                                                          & 51.25                                                                           \\
\multicolumn{1}{|c|}{14}                         & \multicolumn{1}{c|}{81.35}                                                                        & \multicolumn{1}{c|}{Mistral-Medium-2312 (Prompt)}         & 80.55                        & 92                         & 84                         & \multicolumn{1}{c|}{78.5}                                                                          & 65.88                        & 76                         & 82                         & \multicolumn{1}{c|}{70}                                                                            & 88.33                                                                           \\
\multicolumn{1}{|c|}{15}                         & \multicolumn{1}{c|}{80.53}                                                                        & \multicolumn{1}{c|}{GPT-4o-2024-05-13 (Prompt)}           & 85.09                        & 84                         & 78.5                       & \multicolumn{1}{c|}{61}                                                                            & 90                           & 78                         & 70                         & \multicolumn{1}{c|}{72.5}                                                                          & 82.5                                                                            \\
\multicolumn{1}{|c|}{16}                         & \multicolumn{1}{c|}{80.47}                                                                        & \multicolumn{1}{c|}{Functionary-Medium-v2.4 (FC)}         & 79.45                        & 90.5                       & 87.5                       & \multicolumn{1}{c|}{85}                                                                            & 68.82                        & 84                         & 80                         & \multicolumn{1}{c|}{70}                                                                            & 74.17                                                                           \\
\multicolumn{1}{|c|}{17}                         & \multicolumn{1}{c|}{80.35}                                                                        & \multicolumn{1}{c|}{Gemini-1.5-Flash-Preview-0514 (FC)}   & 80.91                        & 93.5                       & 78                         & \multicolumn{1}{c|}{73.5}                                                                          & 81.76                        & 90                         & 54                         & \multicolumn{1}{c|}{72.5}                                                                          & 79.58                                                                           \\
\multicolumn{1}{|c|}{18}                         & \multicolumn{1}{c|}{80.35}                                                                        & \multicolumn{1}{c|}{Command-R-Plus (Prompt) (Optimized)}  & 82.91                        & 88.5                       & 81                         & \multicolumn{1}{c|}{82}                                                                            & 92.94                        & 90                         & 84                         & \multicolumn{1}{c|}{80}                                                                            & 54.17                                                                           \\
\multicolumn{1}{|c|}{19}                         & \multicolumn{1}{c|}{80.29}                                                                        & \multicolumn{1}{c|}{Command-R-Plus (Prompt) (Original)}   & 82.55                        & 90                         & 80                         & \multicolumn{1}{c|}{83}                                                                            & 92.94                        & 88                         & 84                         & \multicolumn{1}{c|}{80}                                                                            & 53.75                                                                           \\
\multicolumn{1}{|c|}{20}                         & \multicolumn{1}{c|}{79.94}                                                                        & \multicolumn{1}{c|}{Functionary-Small-v2.4 (FC)}          & 82.18                        & 88.5                       & 82                         & \multicolumn{1}{c|}{81.5}                                                                          & 78.24                        & 82                         & 80                         & \multicolumn{1}{c|}{65}                                                                            & 67.92                                                                           \\
\multicolumn{1}{|c|}{21}                         & \multicolumn{1}{c|}{79.76}                                                                        & \multicolumn{1}{c|}{Command-R-Plus (FC) (Optimized)}      & 79.09                        & 91                         & 88                         & \multicolumn{1}{c|}{82.5}                                                                          & 81.18                        & 86                         & 74                         & \multicolumn{1}{c|}{67.5}                                                                          & 63.75                                                                           \\
\multicolumn{1}{|c|}{22}                         & \multicolumn{1}{c|}{77.47}                                                                        & \multicolumn{1}{c|}{Claude-3-Opus-0229 (FC)}              & 82.73                        & 91.5                       & 58                         & \multicolumn{1}{c|}{60}                                                                            & 90.59                        & 94                         & 38                         & \multicolumn{1}{c|}{62.5}                                                                          & 82.5                                                                            \\
\multicolumn{1}{|c|}{23}                         & \multicolumn{1}{c|}{76.47}                                                                        & \multicolumn{1}{c|}{Claude-instant-1.2 (Prompt)}          & 79.82                        & 85.5                       & 83                         & \multicolumn{1}{c|}{67.5}                                                                          & 84.71                        & 80                         & 82                         & \multicolumn{1}{c|}{65}                                                                            & 57.5                                                                            \\
\rowcolor[HTML]{DAE8FC} 
\multicolumn{1}{|c|}{\cellcolor[HTML]{DAE8FC}24} & \multicolumn{1}{c|}{\cellcolor[HTML]{DAE8FC}74.41}                                                & \multicolumn{1}{c|}{\cellcolor[HTML]{DAE8FC}xLAM-1B (FC)} & 75.09                        & 80.5                       & 76.5                       & \multicolumn{1}{c|}{\cellcolor[HTML]{DAE8FC}63}                                                    & 79.41                        & 80                         & 78                         & \multicolumn{1}{c|}{\cellcolor[HTML]{DAE8FC}62.5}                                                  & 72.08                                                                           \\
\multicolumn{1}{|c|}{25}                         & \multicolumn{1}{c|}{74.29}                                                                        & \multicolumn{1}{c|}{Claude-3-Haiku-0207 (Prompt)}         & 84.91                        & 91.5                       & 84.5                       & \multicolumn{1}{c|}{55.5}                                                                          & 92.94                        & 94                         & 70                         & \multicolumn{1}{c|}{25}                                                                            & 34.58                                                                           \\
\multicolumn{1}{|c|}{26}                         & \multicolumn{1}{c|}{71.41}                                                                        & \multicolumn{1}{c|}{Claude-2.1 (Prompt)}                  & 80.18                        & 76                         & 55.5                       & \multicolumn{1}{c|}{52.5}                                                                          & 71.18                        & 84                         & 46                         & \multicolumn{1}{c|}{47.5}                                                                          & 83.33                                                                           \\
\multicolumn{1}{|c|}{27}                         & \multicolumn{1}{c|}{70.94}                                                                        & \multicolumn{1}{c|}{Command-R-Plus (FC) (Original)}       & 74.91                        & 90                         & 82                         & \multicolumn{1}{c|}{76.5}                                                                          & 81.76                        & 88                         & 68                         & \multicolumn{1}{c|}{55}                                                                            & 24.17                                                                           \\
\multicolumn{1}{|c|}{28}                         & \multicolumn{1}{c|}{68.76}                                                                        & \multicolumn{1}{c|}{Mistral-large-2402 (FC Auto)}         & 66.91                        & 94.5                       & 25.5                       & \multicolumn{1}{c|}{72}                                                                            & 83.53                        & 96                         & 8                          & \multicolumn{1}{c|}{52.5}                                                                          & 84.17                                                                           \\
\multicolumn{1}{|c|}{29}                         & \multicolumn{1}{c|}{67}                                                                           & \multicolumn{1}{c|}{Gemini-1.0-Pro-001 (FC)}              & 79.09                        & 92.5                       & 30                         & \multicolumn{1}{c|}{25.5}                                                                          & 86.47                        & 84                         & 44                         & \multicolumn{1}{c|}{12.5}                                                                          & 80                                                                              \\
\multicolumn{1}{|c|}{30}                         & \multicolumn{1}{c|}{65.88}                                                                        & \multicolumn{1}{c|}{DBRX-Instruct (Prompt)}               & 64                           & 71.5                       & 72                         & \multicolumn{1}{c|}{59}                                                                            & 71.18                        & 86                         & 80                         & \multicolumn{1}{c|}{62.5}                                                                          & 55.83                                                                           \\
\multicolumn{1}{|c|}{31}                         & \multicolumn{1}{c|}{65.18}                                                                        & \multicolumn{1}{c|}{snowflake-instruct (Prompt)}          & 62.36                        & 69                         & 59                         & \multicolumn{1}{c|}{54}                                                                            & 87.65                        & 86                         & 74                         & \multicolumn{1}{c|}{72.5}                                                                          & 59.58                                                                           \\
\multicolumn{1}{|c|}{32}                         & \multicolumn{1}{c|}{64.35}                                                                        & \multicolumn{1}{c|}{Mistral-large-2402 (FC Any)}          & 81.45                        & 93.5                       & 31.5                       & \multicolumn{1}{c|}{79.5}                                                                          & 94.71                        & 92                         & 8                          & \multicolumn{1}{c|}{65}                                                                            & 0                                                                               \\
\multicolumn{1}{|c|}{33}                         & \multicolumn{1}{c|}{63.88}                                                                        & \multicolumn{1}{c|}{GPT-3.5-Turbo-0125 (FC)}              & 61.45                        & 66                         & 90.5                       & \multicolumn{1}{c|}{81}                                                                            & 93.53                        & 80                         & 82                         & \multicolumn{1}{c|}{70}                                                                            & 2.08                                                                            \\ \hline
\multicolumn{12}{|c|}{... ...}                                                                                                                                                                                                                                                                                                                                                                                                                                                                                                                                                                                                                                                                 \\ \hline
\multicolumn{1}{|c|}{43}                         & \multicolumn{1}{c|}{43.71}                                                                        & \multicolumn{1}{c|}{Gemma-7b-it (Prompt)}                 & 42.18                        & 48                         & 30                         & \multicolumn{1}{c|}{44}                                                                            & 30                           & 32                         & 40                         & \multicolumn{1}{c|}{25}                                                                            & 70.83                                                                           \\
\multicolumn{1}{|c|}{44}                         & \multicolumn{1}{c|}{40.76}                                                                        & \multicolumn{1}{c|}{Mistral-Small-2402 (Prompt)}          & 5.82                         & 68                         & 79                         & \multicolumn{1}{c|}{68.5}                                                                          & 34.12                        & 20                         & 68                         & \multicolumn{1}{c|}{30}                                                                            & 98.33                                                                           \\
\multicolumn{1}{|c|}{45}                         & \multicolumn{1}{c|}{40.41}                                                                        & \multicolumn{1}{c|}{Deepseek-v1.5 (Prompt)}               & 39.27                        & 49                         & 37                         & \multicolumn{1}{c|}{28.5}                                                                          & 37.06                        & 38                         & 36                         & \multicolumn{1}{c|}{12.5}                                                                          & 57.08                                                                           \\
\multicolumn{1}{|c|}{46}                         & \multicolumn{1}{c|}{23.71}                                                                        & \multicolumn{1}{c|}{Mistral-small-2402 (FC Auto)}         & 2                            & 25.5                       & 3                          & \multicolumn{1}{c|}{3}                                                                             & 56.47                        & 70                         & 6                          & \multicolumn{1}{c|}{5}                                                                             & 99.58                                                                           \\ \hline
\end{tabular}
}
\end{table}

%% file: app-1.tex
\section{Dataset Documentation and Accessibility}

\subsection{Dataset Documentation and Intended Uses}

The dataset generated using the APIGen framework is intended for training and evaluating function-calling agents. The dataset consists of diverse query-answer pairs, where the answers are verified function calls that could address the requested query with provided APIs. The APIs and function calls are in a standardized JSON format, as demonstrated in the main paper Fig. 3. More details of the format and examples are available in Appendix \ref{app:json-data-format}. The dataset covers a wide range of API categories and includes various query styles, such as simple, multiple, parallel, and parallel multiple function calls, as introduced in \cite{bfcl}.

\textbf{Hosting, Licensing, and Maintenance Plan.}
The dataset currently can be viewed and downloaded from our project homepage \footnote{\url{https://apigen-pipeline.github.io/}} or via Huggingface \footnote{\url{https://huggingface.co/datasets/Salesforce/xlam-function-calling-60k}}.
All datasets are licensed under the Creative Commons Attribution 4.0 License (CC BY).
We also plan to open-source the trained models on Huggingface once after the company's legal approval. 
As for maintenance, we have established a long-term plan to keep the datasets up-to-date, correct any potential issues, and provide support to users. We also aim to expand these datasets further based on new advances in the field, thus continually promoting progress in the field of function-calling agent training.

\textbf{Author Responsibility Statement.}
As the authors, we hereby affirm that we bear full responsibility for the datasets provided in this submission. We confirm that to the best of our knowledge, no rights are violated in the collection, distribution, and use of these datasets. 

\subsection{JSON Data Format and Examples}
\label{app:json-data-format}
This JSON data format is used to represent a query along with the available tools and the corresponding answers. Here's a description of the format:

\subsubsection{Dataset Structure}

The JSON data structure comprises three main keys: \texttt{query}, a string representing the problem statement; \texttt{tools}, an array of tools each defined by properties such as \texttt{name}, \texttt{description}, and \texttt{parameters} that further describe each tool's required and optional parameters with their types and descriptions; and \texttt{answers}, an array detailing responses with the tool used (\texttt{name}) and the arguments provided (\texttt{arguments}) for each answer, thereby aligning tools with their respective query intentions.
The detailed description of each data point's entries is as follows.

\begin{itemize}[leftmargin=10pt]
  \item \texttt{query} (string): The query or problem statement.
  \item \texttt{tools} (array): An array of available tools that can be used to solve the query.
  \begin{description}
    \item Each tool is represented as an object with the following properties:
    \begin{itemize}
      \item \texttt{name} (string): The name of the tool.
      \item \texttt{description} (string): A brief description of what the tool does.
      \item \texttt{parameters} (object): An object representing the parameters required by the tool.
      \begin{itemize}
        \item Each parameter is represented as a key-value pair, where the key is the parameter name and the value is an object with the following properties:
        \begin{itemize}
          \item \texttt{type} (string): The data type of the parameter (e.g., "integer", "float", "array").
          \item \texttt{description} (string): A brief description of the parameter.
          \item \texttt{required} (boolean): Indicates whether the parameter is required or optional.
        \end{itemize}
      \end{itemize}
    \end{itemize}
  \end{description}
  \item \texttt{answers} (array): An array of answers corresponding to the query.
  \begin{itemize}
    \item Each answer is represented as an object with the following properties:
    \begin{itemize}
      \item \texttt{name} (string): The name of the tool used to generate the answer.
      \item \texttt{arguments} (object): An object representing the arguments passed to the tool to generate the answer.
      \begin{itemize}
        \item Each argument is represented as a key-value pair, where the key is the parameter name and the value is the corresponding value.
      \end{itemize}
    \end{itemize}
  \end{itemize}
\end{itemize}

\subsubsection{Example Data}

Here's an example JSON data for the simplest scenario.
\begin{lstlisting}[language=json]
{
  "query": "What is the weather in Palo Alto?",
  "tools": [
    {
      "name": "weather_api.get_current_weather",
      "description": "Retrieves the current weather conditions for a specified location.",
      "parameters": {
        "location": {
          "type": "string",
          "description": "The name of the city or geographic location.",
          "required": true
        },
        "units": {
          "type": "string",
          "description": "The units for temperature measurement (e.g., 'Celsius', 'Fahrenheit').",
          "required": false
        }
      }
    }
  ],
  "answers": [
    {
      "name": "weather_api.get_current_weather",
      "arguments": {
        "location": "Palo Alto",
        "units": "Celsius"
      }
    }
  ]
}
\end{lstlisting}

In this example, the query asks about the current weather in Palo Alto.
The tools array contains a single entry for \texttt{weather\_api.get\_current\_weather}, describing the tool used to retrieve weather data, including parameters for location and units.
The answers array lists the specific API call made with the location set as \texttt{"Palo Alto"} and units as \texttt{"Celsius"}.

Here's an example JSON data for the parallel function-calling category, i.e., the user's query contains multiple intentions and the answers contain multiple parallel tool calls:

\begin{lstlisting}[language=json]
{
  "query": "Find the sum of all the multiples of 3 and 5 between 1 and 1000. Also find the product of the first five prime numbers.",
  "tools": [
    {
      "name": "math_toolkit.sum_of_multiples",
      "description": "Find the sum of all multiples of specified numbers within a specified range.",
      "parameters": {
        "lower_limit": {
          "type": "integer",
          "description": "The start of the range (inclusive).",
          "required": true
        },
        "upper_limit": {
          "type": "integer",
          "description": "The end of the range (inclusive).",
          "required": true
        },
        "multiples": {
          "type": "array",
          "description": "The numbers to find multiples of.",
          "required": true
        }
      }
    },
    {
      "name": "math_toolkit.product_of_primes",
      "description": "Find the product of the first n prime numbers.",
      "parameters": {
        "count": {
          "type": "integer",
          "description": "The number of prime numbers to multiply together.",
          "required": true
        }
      }
    }
  ],
  "answers": [
    {
      "name": "math_toolkit.sum_of_multiples",
      "arguments": {
        "lower_limit": 1,
        "upper_limit": 1000,
        "multiples": [3, 5]
      }
    },
    {
      "name": "math_toolkit.product_of_primes",
      "arguments": {
        "count": 5
      }
    }
  ]
}
\end{lstlisting}

In this example, the query asks to find the sum of multiples of 3 and 5 between 1 and 1000, and also find the product of the first five prime numbers. The available tools are \texttt{math\_toolkit.sum\_of\_multiples} and \texttt{math\_toolkit.product\_of\_primes}, along with their parameter descriptions. The \texttt{answers} array provides the specific tool and arguments used to generate each answer.

\subsection{Human Evaluation of Dataset Quality}

To ensure that the three-stage verification process employed by APIGen produces a high-quality dataset, we conduct a human evaluation on a sample of the generated data. We engage three human evaluators to manually inspect a total of 600 samples from our released dataset. The evaluators assess the quality of each sample based on factors such as the accuracy of parameter values and the appropriateness of the number of API calls.

The results of the human evaluation reveal that only 28 out of the 600 inspected samples have minor issues, such as inaccurate parameter values or more API calls than expected. This means that the majority of the data, approximately 95.3\%, are of very high quality. The high quality of the dataset can be attributed to the format and execution checkers implemented in the APIGen pipeline.

The format checker ensures that the generated data adheres to the specified JSON format and contains all the necessary fields. This step helps to filter out poorly formatted or incomplete data points. The execution checker, on the other hand, executes the generated function calls against the appropriate backend and verifies their successful execution. By providing real execution results, the execution checker plays a crucial role in filtering out cases that might be difficult to identify by an LLM-based semantic checker alone.

The combination of these two checkers, along with the final semantic checker, creates a robust verification process that effectively filters out low-quality data points. The human evaluation results confirm the effectiveness of this approach, demonstrating that APIGen is capable of generating high-quality datasets for training function-calling agents.

%% file: app-2.tex
\section{Dataset Generation and Experiment Details}
\subsection{Generator LLM Prompt}
\begin{tcolorbox}[title=Example Prompt for the Generator to Generate Parallel Function-Calling Data]
\begin{lstlisting}[language=json2]
"""
You are a data labeler. The responsibility for you is to generate a set of diverse queries and corresponding answers for the given functions in JSON format.

Construct queries and answers that exemplifies how to use these functions in a practical scenario. Include in each query specific, plausible values for each parameter. For instance, if the function requires a date, use a typical and reasonable date.

Ensure the query:
- Is clear and concise
- Contain multiple parallel queries in natural language for the given functions, they could use either the same function with different arguments or different functions
- Demonstrates typical use cases
- Includes all necessary parameters in a meaningful way. For numerical parameters, it could be either numerals or words
- Across a variety level of difficulties, ranging from beginner and advanced use cases
- The corresponding result's parameter types and ranges match with the functions descriptions.

Ensure the answer:
- Is a list of function calls in JSON format.
- The length of the answer list should be equal to the number of requests in the query
- Can solve all the requests in the query effectively

Here are examples of queries and corresponding answers for similar functions:
{examples}

Note that the query could be interpreted as a combination of several independent requests.

Based on these examples and the above instructions, generate {number} diverse query and answer pairs for the functions `{func_name}`. 
The detailed functions description is as follows: 
{func_desc}

{format_inst}

Now please generate {number} diverse query and answer pairs following the above format.
"""
\end{lstlisting}
\end{tcolorbox}
The template provided outlines the prompt for an LLM to generate datasets as data labelers, emphasizing the diversity of query types and complexity to ensure thorough coverage of potential real-world applications. It specifies the importance of generating clear, concise queries and precisely formatted JSON responses. Sampled data, used to populate the \texttt{examples} field, and API information, filling the \texttt{func\_name} and \texttt{func\_desc} fields, enable a structured approach to dataset generation.
The \texttt{format\_inst} specifies the enforced JSON output format, as shown below.

\begin{tcolorbox}[title=Example Format Instruction to Generate Parallel Function-Calling Data]
\begin{lstlisting}[language=json2]
The output MUST strictly adhere to the following JSON format, and NO other text MUST be included:
```
[
  {
    "query": "The generated query.",
    "answers": [
      {
        "name": "api_name",
        "arguments": {
          "arg_name": "value",
          ... (more arguments as required)
        }
      },
      ... (more API calls as required)
    ]
  }
]
```
\end{lstlisting}
\end{tcolorbox}

The enforced JSON output format facilitates efficient data extraction and cost-effective generation. By requesting multiple query-answer pairs in a single inference with the \texttt{number} field—referred to here as a "batching" technique—token usage and costs are significantly reduced.

\subsection{Semantic Checker LLM Prompt}
We prompted another LLM as the semantic checker to evaluate whether the execution results and the tool calls align with the user query. We could use multiple LLMs with different prompts as checkers here to increase the credibility of this verification stage. We provide one example prompt as follows.

\begin{tcolorbox}[title=Example Prompt for the Semantic Checker to Verify the Data]
\begin{lstlisting}[language=json2]
"""
As a data quality evaluator, you must assess the alignment between a user query, corresponding function calls, and their execution results. 
These function calls and results are generated by other models, and your task is to ensure these results accurately reflect the user's intentions.

Do not pass if:

1. The function call does not align with the query's objective, or the input arguments appear incorrect.
2. The function call and arguments are not properly chosen from the available functions.
3. The number of function calls does not correspond to the user's intentions.
4. The execution results are irrelevant and do not match the function's purpose.
5. The execution results contain errors or reflect that the function calls were not executed successfully.

Given Information:
- All Available Functions: 
  {func_desc}

- User Query: {query}

- Generated Function Calls: {func_call}

- Execution Results: {execution_result}

Note: The query may have multiple intentions. Functions may be placeholders, and execution results may be truncated due to length, which is acceptable and should not cause a failure.
The main decision factor is wheather the function calls accurately reflect the query's intentions and the function descriptions.

Provide your reasoning in the thought section and decide if the data passes (answer yes or no). 
If not passing, concisely explain your reasons in the thought section; otherwise, leave this section blank.

Your response MUST strictly adhere to the following JSON format, and NO other text MUST be included.
```
{{
  "thought": "Concisely describe your reasoning here",
  "pass": "yes" or "no"
}}
```
"""
\end{lstlisting}
\end{tcolorbox}

Here, the \texttt{func\_desc} field is the same as the generator, while the \texttt{func\_call} and \texttt{execution\_result} are the key fields to determine whether the generated data successfully address the \texttt{query}'s intention.
We also enforce the model to output a JSON-formatted string, and then extract whether we should give a pass to this data point.

\subsection{Model Training}

We train two function-calling models of different sizes, xLAM-1B (FC) and xLAM-7B (FC), using the dataset generated by APIGen. The training pipeline mainly follows the AgentOhana paper \cite{zhang2024agentohana}. We use 8 NVIDIA A100 40GB GPUs for training both models.

Since the Berkeley Function-Calling Benchmark \cite{bfcl} contains a relevance detection category, which evaluates a model's ability to distinguish non-relevant queries and tools, we extend APIGen to generate relevance detection data points from the generated datasets. These data points cover two types of scenarios: 

\begin{itemize}[leftmargin=10pt]
    \item The provided tools cannot solve the query (e.g., query: "I want to know the weather in Palo Alto on Dec 25, 2023," provided tool: \texttt{get\_house\_price(city)}).
    \item The provided tools are missing key arguments to solve the query (e.g., query: "I want to know the weather in Palo Alto on Dec 25, 2023," provided tool: \texttt{get\_weather(city)}).
\end{itemize}

In both cases, the correct output is an empty tool call or a concise explanation indicating that the model should refuse to answer due to insufficient or irrelevant information.

We create 8,000 such data points from the collected dataset by 1) randomly discarding some tools that will be called in the answer or 2) randomly dropping some required parameters that were used in the generated tool calls. Then we relabel the answer to be an empty tool call or with a concise explanation.
By incorporating relevance detection data points into our training datasets, we can enhance the model's performance in determining when the provided tools are not suitable for addressing a given query. This enables the training of agents that can effectively assess the relevance of the available tools and respond appropriately, either by utilizing the relevant tools or by refraining from answering when the necessary information is lacking.

When training the model, we fill in the sampled query and available tools to the training prompt template, and then ask the model to predict the corresponding tool calls in specified JSON format. The training prompt template is as follows:
\begin{tcolorbox}[title=Model Training Prompt]
\begin{lstlisting}[language=json2]
"""
[BEGIN OF TASK INSTRUCTION]
You are an expert in composing functions. You are given a question and a set of possible functions. 
Based on the question, you will need to make one or more function/tool calls to achieve the purpose. 
If none of the function can be used, point it out and refuse to answer. 
If the given question lacks the parameters required by the function, also point it out.
[END OF TASK INSTRUCTION]

[BEGIN OF AVAILABLE TOOLS]
{func_desc}
[END OF AVAILABLE TOOLS]

[BEGIN OF FORMAT INSTRUCTION]
The output MUST strictly adhere to the following JSON format, and NO other text MUST be included.
The example format is as follows. Please make sure the parameter type is correct. If no function call is needed, please make tool_calls an empty list '[]'
```
{{
  "tool_calls": [
    {{"name": "func_name1", "arguments": {{"argument1": "value1", "argument2": "value2"}}}},
    ... (more tool calls as required)
  ]
}}
```
[END OF FORMAT INSTRUCTION]

[BEGIN OF QUERY]
User Query: {query}
[END OF QUERY]
"""
\end{lstlisting}
\end{tcolorbox}

The training hyperparameters for our models include a learning rate of $5 \times 10^{-6}$, four epochs, and use of the AdamW optimizer. Other settings include a cutoff length of 2048, a per-device batch size of six, two gradient accumulation steps, a cosine learning rate scheduler with 50 warmup steps, and the bfloat16 (BF16) data type. 